\documentclass{article}

\usepackage[preprint]{corl_2021} 
\usepackage{graphicx}
\usepackage{sidecap}
\usepackage{tikz}
\usetikzlibrary{shapes,arrows,positioning,calc}

\title{The Holy Grail of Multi-Robot Planning:\\ Learning to Generate Online-Scalable Solutions from Offline-Optimal Experts}

\author{
  Amanda Prorok\thanks{All authors contributed equally.} \hspace{.5cm}
  Jan Blumenkamp$^*$ \hspace{.5cm}
  Qingbiao Li$^*$ \hspace{.5cm}
  Ryan Kortvelesy$^*$ \hspace{.5cm}
  Zhe Liu$^*$\\
  Department of Computer Science and Technology\\
  University of Cambridge, UK \\
  \texttt{\{asp45,jb2270,ql295,rk627,zl457\}@cam.ac.uk} \\
  \And
  Ethan Stump$^*$\\
  DEVCOM Army Research Laboratory (ARL), Maryland, USA. \\
  \texttt{ethan.a.stump2.civ@mail.mil} \\
  
}

\newcommand{\highlight}[1]{\textbf{\textit{}}}

\begin{document}
\maketitle


\begin{abstract}
    Many multi-robot planning problems are burdened by the curse of dimensionality, which compounds the difficulty of applying solutions to large-scale problem instances. The use of learning-based methods in multi-robot planning holds great promise as it enables us to offload the \textit{online} computational burden of expensive, yet optimal solvers, to an \textit{offline} learning procedure. 
    Simply put, the idea is to train a policy to copy an optimal pattern generated by a small-scale system, and then transfer that policy to much larger systems, in the hope that the learned strategy scales, while maintaining near-optimal performance.
    Yet, a number of issues impede us from leveraging this idea to its full potential. This blue-sky paper elaborates some of the key challenges that remain.
\end{abstract}



\section{Introduction}


Learning-based methods have proven effective at designing robot control policies for an increasing number of tasks~\cite{rajeswaran_Generalization_2017a, tobin_Domain_2017a}. The application of learning-based methods to multi-robot planning has attracted particular attention due to their capability of handling high-dimensional joint state-space representations, by offloading the online computational burden to an offline learning procedure~\cite{tolstaya_Learning_2019a, li_Graph_2020a}. We argue that these developments point to a fundamental approach that combines ideas around the application of learning to optimization and produce a flexible framework that could tackle many hard but important problems in robotics, including multi-agent path planning~\cite{yu_Structure_2013}, area coverage~\cite{schwager_Decentralized_2009, rabban2020improved}, task allocation~\cite{ponda_Distributed_2012a, prorok_Redundant_2019a, zavlanos_distributed_2008a}, formation control~\cite{michael_Distributed_2008a}, and target-tracking~\cite{jung_Cooperative_2006}. In this paper, we motivate this approach and discuss the crucial challenges and research questions.


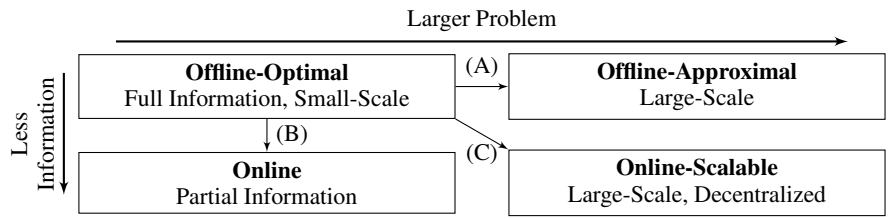
\begin{figure}[bh]
    \centering

    \begin{tikzpicture}[auto, node distance=1cm,>=latex']
        \tikzstyle{every node}=[font=\small]
        \tikzstyle{box}=[minimum width=5cm, draw, align=center]
        \tikzstyle{axis}=[thick]
        
        \node [box] (offline_optimal) {\textbf{Offline-Optimal}\\Full Information, Small-Scale};
        \node [box, below=1.25em of offline_optimal] (online_partial) {\textbf{Online}\\Partial Information};
        \node [box, right=2em of offline_optimal] (offline_approx) {\textbf{Offline-Approximal}\\Large-Scale};
        \node [box, right=2em of online_partial] (online_large_scale) {\textbf{Online-Scalable}\\Large-Scale, Decentralized};

        \draw[->, axis]($(offline_optimal.west)+(0.5,0.6)$) -- node [pos=0.5, above] {Larger Problem}($(offline_approx.east)+(-0.5,0.6)$);
        \draw[->, axis]($(offline_optimal.north)+(-2.7,-0.25)$) -- node [rotate=90, pos=0.5, align=center,above] {Less\\Information}($(online_partial.south)+(-2.7,0.25)$);
                
        \draw[->](offline_optimal) -- node [pos=0.5, above] {(A)}(offline_approx);
        \draw[->](offline_optimal) -- node [pos=0.5, right] {(B)}(online_partial);
        \draw[->](offline_optimal.south east) -- node [pos=0.5, below] {(C)}(online_large_scale.north west);
    \end{tikzpicture}
    
    \caption{\small{Applications of learning to optimization problems. (A) embodies techniques for learning optimization heuristics; (B) embodies techniques for learning to solve POMDPs; (C) is the emerging topic discussed here, embodying techniques for learning to coordinate large systems in real-world applications}}
    \label{fig:taxonomy}
\end{figure}

The ideas here sit within a larger landscape of the application of learning to the solution of optimization problems. Consider Figure~\ref{fig:taxonomy}, where we consider how learning is applied to either increase the scale of solvable problems or to increase the ability to deal with practical, partial-information problems. Along the problem scale axis, for example, the operations research community has made use of learned heuristics to solve TSPs~\cite{deudon2018learning}, VRPs~\cite{nazari2018reinforcement}, and general MILPs~\cite{khalil2017learning}. Along the information axis, which includes dealing with POMDPs, techniques such as RL play a major role, as well as ideas such as tuning Monte-Carlo Tree Search~\cite{katt2017learning}, embedding learned components into optimal control frameworks~\cite{richter2014high}, and learning how to bias sampling planners~\cite{liu2020learned}.


Practical multi-robot planning and control builds on the progress along both of these axes: the degrees of freedom and environment complexity increase, while the ability to communicate and coordinate at scale decreases. Traditional \textit{centralized} approaches would use a planning unit to produce coordinated plans that agents use for real-time on-board control; these have the advantage of producing optimal and complete plans in the joint configuration space but true optimality is NP-hard in many cases~\cite{yu_Structure_2013} and they will struggle when communications are degraded and frequent replanning is required. By contrast, \textit{decentralized} approaches reduce the computational overhead~\cite{desaraju:2012} and relax the dependence on centralized units~\cite{vandenberg:2008,wang_Mobile_2020} to deal with challenged communications, but account for purely local objectives and cannot explicitly optimize global objectives (e.g., path efficiency).


What the directions of Figure~\ref{fig:taxonomy} teach us is that success follows from starting with simple problems and using their examples to approach complex ones. This progression from example to application is reminiscent of \textit{Imitation Learning}, and we use this crucial observation to understand how learning can play a role in mitigating the shortcomings of decentralized approaches in solving challenging multi-robot problems. 


Bridging the gap between the qualities of centralized and decentralized approaches, learning-based methods promise to find solutions that \textit{balance optimality and real-world efficiency}. The process of generating data-driven solutions for multi-robot systems, however, cannot directly borrow from single-robot learning methods because \emph{(a)} hidden (unobservable) information about other robots must be incorporated through learned communication strategies, and \emph{(b)}, although policies are executed locally, the ensuing actions should lead to plans with a performance near to that of coupled systems. This agenda means that we need to address \emph{(i)} how to generate multi-robot training data, \emph{(ii)} how to generate decentralizable policies, and \emph{(iii)} how to transfer these policies to real-world systems.

The following section elaborates these three key challenges and indicates promising directions.

\section{Learning Decentralized Policies by Copying Centralized Experts}
\label{sec:challenges}


Though planning complexity is reduced with a decentralized approach, use of a learning-based approach requires consideration of state-action space coverage, especially since introducing multiple agents means the size of the joint state-action again grows exponentially. This core challenge is the reason why the development of learning-based multi-robot controllers is a nascent field. While a number of learning paradigms have been applied to this topic (e.g., RL~\cite{vinyals2019grandmaster, blumenkamp_2020_adversarial}), this position paper focuses on imitation learning strategies.
The following paragraphs discuss three key topics that are central to the learning process: data generation, communication strategies, and sim-to-real transfer.

\subsection{Experts and Data Generation}


\textbf{\textit{How to generate expert data?}}
The work by~\citet{li_Graph_2020a} shows that it is possible to train decentralized controllers to learn communication and action policies that optimize a global objective by imitating a centralized optimal expert. The former work considered the specific case-study of multi-agent path planning, and used Conflict-Based Search (CBS)~\citep{Sharon15-CBS} to find optimal solutions (i.e., sets of optimal, collision-free paths).
%
Although their results demonstrated unprecedented performance in decentralized systems (i.e., achieving higher than 96\% success rates with single-digit flowtime increases, compared to the expert solution), but observed poor generalization.
Simply training the models through behavior cloning leads to bias and over-fitting, since the performance of the network is intrinsically constrained by the dataset.
Alternative approaches include learning curricula~\cite{bengio2009curriculum} to optimize the usage of the existing training set, or the introduction of data augmentation mechanisms, which allow experts to teach the learner how to recover from past mistakes.


\textbf{\textit{How to augment existing datasets?}}
One of the major limitations of behavior cloning is that it does not learn to recover from failures, and is unable to handle unseen situations~\cite{attia2018global}. For example, if the policy has deviated from the optimal trajectory at one-time step, it will fail in getting back to states seen by the expert, hence, resulting in a cascade of errors. One solution (i.e., DAgger~\citep{ross2011reduction}) is to introduce the expert \textit{during} training to teach the learner how to recover from past mistakes. In~\cite{li_Graph_2020a}, the authors demonstrate the utility of this approach by making use of a novel dataset aggregation method that leverages an online expert to resolve hard cases during training.  Other approaches are to directly extract a policy from training data, such as GAIL \citep{ho2016generative}.
More broadly speaking, with data augmentation, one can produce arbitrary amounts of training data from arbitrary probability distributions to account for a variety of factors, such as roadmap structure, local environment, obstacle density, motion characteristics, and local robot configurations. Such carefully controlled distributions enable us to introduce different levels of local coordination difficulties and generate the most challenging instances at each training stage, inherently achieving a form of curriculum learning. In addition, data augmentation allows us to understand the ability boundary of the trained model, to analyze the correlation between different factors, and to find identify factors that have the strongest effect on the system performance. 

\subsection{Communication Strategies for Decentralized Control}

\textbf{\textit{What, how and when to send information?}} 
While effective communication is key to decentralized control, it is far from obvious \emph{what information is crucial to the task, and what must be shared among agents.} This question differs from problem to problem and the optimal strategy is often unknown. Hand-engineered coordination strategies often fail to deliver the desired performance, and despite ongoing progress in this domain, they still require substantial design effort.
Recent work has shown the promise of Graph Neural Networks (GNNs) to learn explicit communication strategies that enable complex multi-agent coordination~\cite{modgnn,khan_2020, tolstaya_Learning_2019a, li_Graph_2020a}. 
In the context of multi-robot systems, individual robots are modeled as nodes, the communication links between them as edges, and the internal state of each robot as graph signals. By sending messages over the communication links, each robot in the graph indirectly receives access to the global state. The key attribute of GNNs is that they compress data as it flows through the communication graph. 
In effect, this compresses the global state, affording agents access to global data without inundating them with the entire raw global state. Since compression is performed on local networks (with parameters that can be shared across the entire graph), GNNs are able to compress previously unseen global states. In the process of learning how to compress the global state, GNNs also learn which elements of the signal are the most important, and discard the irrelevant information~\citep{modgnn}. This produces a non-injective mapping from global states to latent states, where similar global states `overlap', further improving generalization.


\textbf{\textit{Are all messages equally important?}}
Unfortunately, if communication happens concurrently and equivalently among many neighboring robots, it is likely to cause redundant information, burden the computational capacity and adversely affect overall team performance. Hence, new approaches towards \textit{communication-aware planning} are required.
%
A potential approach is to introduce \textit{attention mechanisms} to actively measure the relative importance of messages (and their senders). Attention mechanisms have been actively studied and widely adopted in various learning-based models~\cite{vaswani2017attention}, which can be viewed as dynamically amplifying or reducing the weights of features based on their relative importance computed by a given mechanism.
Hence, the network can be trained to focus on task-relevant parts of the graph~\cite{velivckovic2017graph}. Learning attention over static graphs has shown to be efficient. 
Liu et al.~\cite{liu2020when2com} developed a learning-based communication model that constructs the communication group on a static graph to address what to transmit and which agent to communicate to for collaborative perception.
However, its permutation equivariance, time invariance and its practical effectiveness in dynamic multi-agent communication graphs have not yet been verified. 
Recently,~\citet{li_Messageaware_2021} integrated an attention mechanism with a GNN-based communication strategy to allow for \textit{message-dependent attention} in a multi-agent path planning problem. A key-query-like mechanism determines the relative importance of features in the messages received from various neighboring robots. Their results show that it is possible to achieve performance close to that of a coupled centralized expert algorithm, while scaling to problem instances that are $\times 100$ larger than the training instances.

\subsection{Sim-to-Real Transfer}
Expert data is typically generated in a simulation, yet policies trained in simulation often do not generalize to the real world. This is referred to as the~\textit{reality gap}~\cite{jakobi_1995}. 

\textbf{\textit{Why is sim-to-real transfer difficult?}}
Even though simulations have become more realistic and easily accessible over recent years \cite{dosovitskiy_17_carla, coumans_2021_pybullet}, it is computationally infeasible to replicate all aspects of real-world physics in a simulation since the uncertainty and randomness of complex robot-world interactions are difficult to model. Domain randomization is an intuitive solution to this problem, but also makes the task to learn harder than necessary and therefore results in sub-optimal policies.
While the reality gap is a major challenge in computer vision, robotics also deals with the physical interaction with the real world and physical constraints such as inertia, for example in robotic grasping \cite{james_2019_grasping, bousmalis_2018_grasp_domain_adapt}, drone flight \cite{loquercio_2019_drone_rand, kaufmann_2020_drone_acro} or robotic locomotion \cite{nagabandi_2018, tan_2018}.

\textbf{\textit{Why is sim-to-real transfer even more difficult for multi-robot systems?}}
While sim-to-real in the single-robot domain typically deals with robot-world interaction, the multi-robot domain is also concerned with robot-robot interactions. An example of this is a swarm of drones flying closely to each other and turbulence affecting the motions of other drones in the vicinity. We already have established that communication is key to efficient multi-robot interaction, but it is not obvious how such communications are affected by the reality gap.
Multi-robot coordination is typically trained in a synchronous manner, but when deploying these policies to the real-world, decentralized communication is \textit{asynchronous}. Furthermore, randomness such as message dropouts and delays are typically not considered during synchronous training. To the best of our knowledge, no research has been conducted that evaluates those factors and the impact they have on the performance of policies. Decentralization is key to successful multi-agent systems, therefore decentralized mesh communication networks are required to operate multi-robot systems in the real world, which may pose additional challenges to the sim-to-real transfer.
Lastly, during cooperative training it is typically assumed that all agents are being truthful about their communications, but faulty and malicious agents can be part of the real world and cause additional problems \cite{blumenkamp_2020_adversarial, mitchell_2020_gp}.

\textbf{\textit{How can we close the reality gap?}} 
We see a few possible avenues to tackle the sim-to-real transfer for multi-robot communication. Domain randomization facilitates the process of making the real-world a permutation of the training environment, and likely improves performance, potentially even against faulty agents and adversarial attacks \cite{blumenkamp_2020_adversarial,mitchell_2020_gp}, yet leading to sub-optimal policies. More realistic (network) simulations \cite{calvo_2021_rosnetsim} are always helpful, but also costly alternatives. Methods such as \textit{sim-to-real via real-to-sim} \cite{zhang_2019_vr_goggles} or training agents in the real-world in a \textit{mixed reality} setting \cite{mitchell_2020_mixed_reality} and federated, decentralized learning where individual robots collect data and use it to update a local model that is then aggregated into a global model can benefit the sim-to-real transfer \cite{mcmahan_2017_federated, wang_2021_coding_federated}.


	


\section{Future Avenues}
\label{sec:conclusion}

The sections above lay out the challenges entailed by the described approach. Yet, this begs the following two questions:

\textbf{\textit{Is imitation learning the right paradigm?}} There are two main approaches to training a controller for a multi-robot system: imitation learning (e.g., \cite{le2017coordinated}) and reinforcement learning (e.g., \cite{lowe_MultiAgent_2017a}). The most obvious benefit to RL is that it does not require an expert algorithm, as it simply optimizes a reward. However, the reward function requires careful consideration to guarantee that the learned controller does not exploit it by using unsafe or inappropriate actions. Conversely, IL is often biased around regions that can be reached by the expert and, consequently, if the controller ever finds itself in a previously unseen situation, it might exhibit unpredictable behavior. Finally, IL is inherently limited by the expert algorithm. As such, possible future directions should explore the combination of both IL and RL (e.g., \cite{vinyals2019grandmaster}) in the context of decentralized multi-robot systems.

\textbf{\textit{Is it possible to learn small-scale coordination patterns for large-scale systems?}}
%
Ideally, we hope that controllers trained on only a few robots (which not only facilitates data generation, but also accelerates the training process), can then be deployed on large-scale systems with hundreds and even thousands of robots. 
Achieving this expectation may be within our reach. A recent example can be found in \cite{Damani2021_PRIMAL}, where the local coordination behaviors and conventions learned in a partially observable world successfully scales up to $2048$ mobile robots in crowded and highly-structured environments. 
In \cite{li_Messageaware_2021}, a promising demonstration shows that the policy trained in $20\times 20$ maps with only $10$ robots obtains a success rate above $80\%$ in $200\times 200$ maps with $1000$ robots, and more impressively, the learned policy only spends $\frac{1}{30}$ computation time compared to the centralized expert. 
Overall, these preliminary results give us confidence that we should continue leveraging methods, such as IL, to distill offline-optimal algorithms to online-scalable controllers.





\acknowledgments{
We gratefully acknowledge the support of ARL grant DCIST CRA W911NF-17-2-0181, Engineering and Physical Sciences Research Council (grant EP/S015493/1), and European Research Council (ERC) Project 949940 (gAIa).  
}



\bibliography{bib,bib_jan,bib_liu}  

\begin{thebibliography}{52}
\providecommand{\natexlab}[1]{#1}
\providecommand{\url}[1]{\texttt{#1}}
\expandafter\ifx\csname urlstyle\endcsname\relax
  \providecommand{\doi}[1]{doi: #1}\else
  \providecommand{\doi}{doi: \begingroup \urlstyle{rm}\Url}\fi

\bibitem[Rajeswaran et~al.(2017)Rajeswaran, Lowrey, Todorov, and
  Kakade]{rajeswaran_Generalization_2017a}
A.~Rajeswaran, K.~Lowrey, E.~V. Todorov, and S.~M. Kakade.
\newblock Towards generalization and simplicity in continuous control.
\newblock In I.~Guyon, U.~V. Luxburg, S.~Bengio, H.~Wallach, R.~Fergus,
  S.~Vishwanathan, and R.~Garnett, editors, \emph{Advances in {Neural}
  {Information} {Processing} {Systems} ({NIPS})}, volume~30, pages 6550--6561.
  Curran Associates, Inc., 2017.

\bibitem[Tobin et~al.(2017)Tobin, Fong, Ray, Schneider, Zaremba, and
  Abbeel]{tobin_Domain_2017a}
J.~Tobin, R.~Fong, A.~Ray, J.~Schneider, W.~Zaremba, and P.~Abbeel.
\newblock Domain randomization for transferring deep neural networks from
  simulation to the real world.
\newblock In \emph{2017 {IEEE}/{RSJ} {International} {Conference} on
  {Intelligent} {Robots} and {Systems} ({IROS})}, pages 23--30, Sept. 2017.
\newblock \doi{10.1109/IROS.2017.8202133}.

\bibitem[Tolstaya et~al.(2019)Tolstaya, Gama, Paulos, Pappas, Kumar, and
  Ribeiro]{tolstaya_Learning_2019a}
E.~Tolstaya, F.~Gama, J.~Paulos, G.~Pappas, V.~Kumar, and A.~Ribeiro.
\newblock Learning {Decentralized} {Controllers} for {Robot} {Swarms} with
  {Graph} {Neural} {Networks}.
\newblock \emph{arXiv:1903.10527 [cs]}, Mar. 2019.
\newblock URL \url{http://arxiv.org/abs/1903.10527}.
\newblock arXiv: 1903.10527.

\bibitem[Li et~al.()Li, Gama, Ribeiro, and Prorok]{li_Graph_2020a}
Q.~Li, F.~Gama, A.~Ribeiro, and A.~Prorok.
\newblock Graph neural networks for decentralized multi-robot path planning.
\newblock In \emph{2020 IEEE/RSJ International Conference on Intelligent Robots
  and Systems (IROS)}, pages 11785--11792. IEEE.

\bibitem[Yu and LaValle(2013)]{yu_Structure_2013}
J.~Yu and S.~M. LaValle.
\newblock Structure and intractability of optimal multi-robot path planning on
  graphs.
\newblock In \emph{{AAAI}}, 2013.

\bibitem[Schwager et~al.(2009)Schwager, Rus, and
  Slotine]{schwager_Decentralized_2009}
M.~Schwager, D.~Rus, and J.-J. Slotine.
\newblock Decentralized, adaptive coverage control for networked robots.
\newblock \emph{The International Journal of Robotics Research}, 28\penalty0
  (3):\penalty0 357--375, 2009.

\bibitem[Rabban and Tokekar(2020)]{rabban2020improved}
I.~E. Rabban and P.~Tokekar.
\newblock Improved resilient coverage maximization with multiple robots.
\newblock \emph{arXiv preprint arXiv:2007.02204}, 2020.

\bibitem[Ponda et~al.(2012)Ponda, Johnson, and How]{ponda_Distributed_2012a}
S.~S. Ponda, L.~B. Johnson, and J.~P. How.
\newblock Distributed chance-constrained task allocation for autonomous
  multi-agent teams.
\newblock In \emph{American {Control} {Conference} ({ACC})}, pages 4528--4533,
  2012.

\bibitem[Prorok(2019)]{prorok_Redundant_2019a}
A.~Prorok.
\newblock Redundant {Robot} {Assignment} on {Graphs} with {Uncertain} {Edge}
  {Costs}.
\newblock In N.~Correll, M.~Schwager, and M.~Otte, editors, \emph{Distributed
  {Autonomous} {Robotic} {Systems}}, Springer {Proceedings} in {Advanced}
  {Robotics}, pages 313--327, 2019.
\newblock ISBN 978-3-030-05816-6.

\bibitem[Zavlanos et~al.(2008)Zavlanos, Spesivtsev, and
  Pappas]{zavlanos_distributed_2008a}
M.~M. Zavlanos, L.~Spesivtsev, and G.~J. Pappas.
\newblock A distributed auction algorithm for the assignment problem.
\newblock In \emph{2008 47th {IEEE} {Conference} on {Decision} and {Control}},
  pages 1212--1217, Dec. 2008.
\newblock \doi{10.1109/CDC.2008.4739098}.
\newblock ISSN: 0191-2216.

\bibitem[Michael et~al.(2008)Michael, Zavlanos, Kumar, and
  Pappas]{michael_Distributed_2008a}
N.~Michael, M.~M. Zavlanos, V.~Kumar, and G.~J. Pappas.
\newblock Distributed multi-robot task assignment and formation control.
\newblock In \emph{{IEEE} {International} {Conference} {Robotics} and
  {Automation}}, pages 128--133, 2008.

\bibitem[Jung and Sukhatme(2006)]{jung_Cooperative_2006}
B.~Jung and G.~S. Sukhatme.
\newblock Cooperative multi-robot target tracking.
\newblock In \emph{Distributed autonomous robotic systems 7}, pages 81--90.
  Springer, 2006.

\bibitem[Deudon et~al.(2018)Deudon, Cournut, Lacoste, Adulyasak, and
  Rousseau]{deudon2018learning}
M.~Deudon, P.~Cournut, A.~Lacoste, Y.~Adulyasak, and L.-M. Rousseau.
\newblock Learning heuristics for the tsp by policy gradient.
\newblock In \emph{International conference on the integration of constraint
  programming, artificial intelligence, and operations research}, pages
  170--181. Springer, 2018.

\bibitem[Nazari et~al.(2018)Nazari, Oroojlooy, Snyder, and
  Tak{\'a}{\v{c}}]{nazari2018reinforcement}
M.~Nazari, A.~Oroojlooy, L.~V. Snyder, and M.~Tak{\'a}{\v{c}}.
\newblock Reinforcement learning for solving the vehicle routing problem.
\newblock \emph{arXiv preprint arXiv:1802.04240}, 2018.

\bibitem[Khalil et~al.(2017)Khalil, Dilkina, Nemhauser, Ahmed, and
  Shao]{khalil2017learning}
E.~B. Khalil, B.~Dilkina, G.~L. Nemhauser, S.~Ahmed, and Y.~Shao.
\newblock Learning to run heuristics in tree search.
\newblock In \emph{IJCAI}, pages 659--666, 2017.

\bibitem[Katt et~al.(2017)Katt, Oliehoek, and Amato]{katt2017learning}
S.~Katt, F.~A. Oliehoek, and C.~Amato.
\newblock Learning in pomdps with monte carlo tree search.
\newblock In \emph{International Conference on Machine Learning}, pages
  1819--1827. PMLR, 2017.

\bibitem[Richter et~al.(2014)Richter, Ware, and Roy]{richter2014high}
C.~Richter, J.~Ware, and N.~Roy.
\newblock High-speed autonomous navigation of unknown environments using
  learned probabilities of collision.
\newblock In \emph{2014 IEEE International Conference on Robotics and
  Automation (ICRA)}, pages 6114--6121. IEEE, 2014.

\bibitem[Liu et~al.(2020)Liu, Stadler, and Roy]{liu2020learned}
K.~Liu, M.~Stadler, and N.~Roy.
\newblock Learned sampling distributions for efficient planning in hybrid
  geometric and object-level representations.
\newblock In \emph{2020 IEEE International Conference on Robotics and
  Automation (ICRA)}, pages 9555--9562. IEEE, 2020.

\bibitem[Desaraju and How(2012)]{desaraju:2012}
V.~R. Desaraju and J.~P. How.
\newblock Decentralized path planning for multi-agent teams with complex
  constraints.
\newblock \emph{Autonomous Robots}, 32\penalty0 (4):\penalty0 385--403, 2012.
\newblock Publisher: Springer.

\bibitem[Van~den Berg et~al.(2008)Van~den Berg, Lin, and
  Manocha]{vandenberg:2008}
J.~Van~den Berg, M.~Lin, and D.~Manocha.
\newblock Reciprocal velocity obstacles for real-time multi-agent navigation.
\newblock In \emph{{IEEE} international conference on robotics and automation
  ({ICRA})}, pages 1928--1935, 2008.
\newblock tex.organization: IEEE.

\bibitem[Wang et~al.(2020)Wang, Liu, Li, and Prorok]{wang_Mobile_2020}
B.~Wang, Z.~Liu, Q.~Li, and A.~Prorok.
\newblock Mobile robot path planning in dynamic environments through globally
  guided reinforcement learning.
\newblock \emph{IEEE Robotics and Automation Letters}, 5\penalty0 (4):\penalty0
  6932--6939, 2020.

\bibitem[Vinyals et~al.(2019)Vinyals, Babuschkin, Czarnecki, Mathieu, Dudzik,
  Chung, Choi, Powell, Ewalds, Georgiev, Oh, Horgan, Kroiss, Danihelka, Huang,
  Sifre, Cai, Agapiou, Jaderberg, Vezhnevets, Leblond, Pohlen, Dalibard,
  Budden, Sulsky, Molloy, Paine, Gulcehre, Wang, Pfaff, Wu, Ring, Yogatama,
  Wünsch, McKinney, Smith, Schaul, Lillicrap, Kavukcuoglu, Hassabis, Apps, and
  Silver]{vinyals2019grandmaster}
O.~Vinyals, I.~Babuschkin, W.~M. Czarnecki, M.~Mathieu, A.~Dudzik, J.~Chung,
  D.~H. Choi, R.~Powell, T.~Ewalds, P.~Georgiev, J.~Oh, D.~Horgan, M.~Kroiss,
  I.~Danihelka, A.~Huang, L.~Sifre, T.~Cai, J.~P. Agapiou, M.~Jaderberg, A.~S.
  Vezhnevets, R.~Leblond, T.~Pohlen, V.~Dalibard, D.~Budden, Y.~Sulsky,
  J.~Molloy, T.~L. Paine, C.~Gulcehre, Z.~Wang, T.~Pfaff, Y.~Wu, R.~Ring,
  D.~Yogatama, D.~Wünsch, K.~McKinney, O.~Smith, T.~Schaul, T.~Lillicrap,
  K.~Kavukcuoglu, D.~Hassabis, C.~Apps, and D.~Silver.
\newblock Grandmaster level in starcraft ii using multi-agent reinforcement
  learning.
\newblock In \emph{Nature}, 2019.
\newblock URL \url{https://www.nature.com/articles/s41586-019-1724-z.pdf}.

\bibitem[Blumenkamp and Prorok(2020)]{blumenkamp_2020_adversarial}
J.~Blumenkamp and A.~Prorok.
\newblock The emergence of adversarial communication in multi-agent
  reinforcement learning.
\newblock \emph{Conference on Robot Learning (CoRL)}, 2020.

\bibitem[Sharon et~al.(2015)Sharon, Stern, Felner, and
  Sturtevant]{Sharon15-CBS}
G.~Sharon, R.~Stern, A.~Felner, and N.~R. Sturtevant.
\newblock Conflict-based search for optimal multi-agent pathfinding.
\newblock \emph{Artificial Intelligence}, 219:\penalty0 40--66, 2015.

\bibitem[Bengio et~al.(2009)Bengio, Louradour, Collobert, and
  Weston]{bengio2009curriculum}
Y.~Bengio, J.~Louradour, R.~Collobert, and J.~Weston.
\newblock Curriculum learning.
\newblock In \emph{Proceedings of the 26th annual international conference on
  machine learning}, pages 41--48, 2009.

\bibitem[Attia and Dayan(2018)]{attia2018global}
A.~Attia and S.~Dayan.
\newblock Global overview of imitation learning.
\newblock \emph{arXiv preprint arXiv:1801.06503}, 2018.

\bibitem[Ross et~al.(2011)Ross, Gordon, and Bagnell]{ross2011reduction}
S.~Ross, G.~Gordon, and D.~Bagnell.
\newblock A reduction of imitation learning and structured prediction to
  no-regret online learning.
\newblock In \emph{Proceedings of the fourteenth international conference on
  artificial intelligence and statistics}, pages 627--635. JMLR Workshop and
  Conference Proceedings, 2011.

\bibitem[Ho and Ermon(2016)]{ho2016generative}
J.~Ho and S.~Ermon.
\newblock Generative adversarial imitation learning.
\newblock \emph{Advances in neural information processing systems},
  29:\penalty0 4565--4573, 2016.

\bibitem[Kortvelesy and Prorok(2021)]{modgnn}
R.~Kortvelesy and A.~Prorok.
\newblock Modgnn: Expert policy approximation in multi-agent systems with a
  modular graph neural network architecture.
\newblock \emph{International Conference on Robotics and Automation (ICRA)},
  2021.

\bibitem[Khan et~al.(2020)Khan, Tolstaya, Ribeiro, and Kumar]{khan_2020}
A.~Khan, E.~Tolstaya, A.~Ribeiro, and V.~Kumar.
\newblock Graph policy gradients for large scale robot control.
\newblock In \emph{Conference on Robot Learning}, pages 823--834, 2020.

\bibitem[Vaswani et~al.(2017)Vaswani, Shazeer, Parmar, Uszkoreit, Jones, Gomez,
  Kaiser, and Polosukhin]{vaswani2017attention}
A.~Vaswani, N.~Shazeer, N.~Parmar, J.~Uszkoreit, L.~Jones, A.~N. Gomez,
  {\L}.~Kaiser, and I.~Polosukhin.
\newblock Attention is all you need.
\newblock In \emph{Advances in Neural Information Processing Systems}, pages
  5998--6008, 2017.

\bibitem[Veli{\v{c}}kovi{\'{c}} et~al.(2018)Veli{\v{c}}kovi{\'{c}}, Cucurull,
  Casanova, Romero, Li{\`{o}}, and Bengio]{velivckovic2017graph}
P.~Veli{\v{c}}kovi{\'{c}}, G.~Cucurull, A.~Casanova, A.~Romero, P.~Li{\`{o}},
  and Y.~Bengio.
\newblock {Graph attention networks}.
\newblock \emph{International Conference on Learning Representations}, 2018.

\bibitem[Liu et~al.(2020)Liu, Tian, Glaser, and Kira]{liu2020when2com}
Y.-C. Liu, J.~Tian, N.~Glaser, and Z.~Kira.
\newblock When2com: multi-agent perception via communication graph grouping.
\newblock In \emph{IEEE Conference on Computer Vision and Pattern Recognition},
  pages 4106--4115, 2020.

\bibitem[Li et~al.(2021)Li, Lin, Liu, and Prorok]{li_Messageaware_2021}
Q.~Li, W.~Lin, Z.~Liu, and A.~Prorok.
\newblock Message-aware graph attention networks for large-scale multi-robot
  path planning.
\newblock \emph{IEEE Robotics and Automation Letters}, 6\penalty0 (3):\penalty0
  5533--5540, 2021.

\bibitem[Jakobi et~al.(1995)Jakobi, Husbands, and Harvey]{jakobi_1995}
N.~Jakobi, P.~Husbands, and I.~Harvey.
\newblock Noise and the reality gap: The use of simulation in evolutionary
  robotics.
\newblock In \emph{Advances in Artificial Life}, volume 929. Springer, 1995.

\bibitem[Dosovitskiy et~al.(2017)Dosovitskiy, Ros, Codevilla, Lopez, and
  Koltun]{dosovitskiy_17_carla}
A.~Dosovitskiy, G.~Ros, F.~Codevilla, A.~Lopez, and V.~Koltun.
\newblock {CARLA}: {An} open urban driving simulator.
\newblock In \emph{Proceedings of the 1st Annual Conference on Robot Learning},
  pages 1--16, 2017.

\bibitem[Coumans and Bai(2016--2021)]{coumans_2021_pybullet}
E.~Coumans and Y.~Bai.
\newblock Pybullet, a python module for physics simulation for games, robotics
  and machine learning.
\newblock \url{http://pybullet.org}, 2016--2021.

\bibitem[James et~al.(2019)James, Wohlhart, Kalakrishnan, Kalashnikov, Irpan,
  Ibarz, Levine, Hadsell, and Bousmalis]{james_2019_grasping}
S.~James, P.~Wohlhart, M.~Kalakrishnan, D.~Kalashnikov, A.~Irpan, J.~Ibarz,
  S.~Levine, R.~Hadsell, and K.~Bousmalis.
\newblock Sim-to-real via sim-to-sim: Data-efficient robotic grasping via
  randomized-to-canonical adaptation networks.
\newblock In \emph{2019 IEEE/CVF Conference on Computer Vision and Pattern
  Recognition (CVPR)}, pages 12619--12629, Los Alamitos, CA, USA, jun 2019.
  IEEE Computer Society.
\newblock \doi{10.1109/CVPR.2019.01291}.
\newblock URL
  \url{https://doi.ieeecomputersociety.org/10.1109/CVPR.2019.01291}.

\bibitem[Bousmalis et~al.(2018)Bousmalis, Irpan, Wohlhart, Bai, Kelcey,
  Kalakrishnan, Downs, Ibarz, Pastor, Konolige,
  et~al.]{bousmalis_2018_grasp_domain_adapt}
K.~Bousmalis, A.~Irpan, P.~Wohlhart, Y.~Bai, M.~Kelcey, M.~Kalakrishnan,
  L.~Downs, J.~Ibarz, P.~Pastor, K.~Konolige, et~al.
\newblock Using simulation and domain adaptation to improve efficiency of deep
  robotic grasping.
\newblock In \emph{2018 IEEE international conference on robotics and
  automation (ICRA)}, pages 4243--4250. IEEE, 2018.

\bibitem[Loquercio et~al.(2019)Loquercio, Kaufmann, Ranftl, Dosovitskiy,
  Koltun, and Scaramuzza]{loquercio_2019_drone_rand}
A.~Loquercio, E.~Kaufmann, R.~Ranftl, A.~Dosovitskiy, V.~Koltun, and
  D.~Scaramuzza.
\newblock Deep drone racing: From simulation to reality with domain
  randomization.
\newblock \emph{IEEE Transactions on Robotics}, 36\penalty0 (1):\penalty0
  1--14, 2019.

\bibitem[Kaufmann et~al.(2020)Kaufmann, Loquercio, Ranftl, M{\"u}ller, Koltun,
  and Scaramuzza]{kaufmann_2020_drone_acro}
E.~Kaufmann, A.~Loquercio, R.~Ranftl, M.~M{\"u}ller, V.~Koltun, and
  D.~Scaramuzza.
\newblock Deep drone acrobatics.
\newblock In \emph{Proceedings of Robotics: Science and Systems}, Corvalis,
  Oregon, USA, July 2020.
\newblock \doi{10.15607/RSS.2020.XVI.040}.

\bibitem[Nagabandi et~al.(2018)Nagabandi, Clavera, Liu, Fearing, Abbeel,
  Levine, and Finn]{nagabandi_2018}
A.~Nagabandi, I.~Clavera, S.~Liu, R.~S. Fearing, P.~Abbeel, S.~Levine, and
  C.~Finn.
\newblock Learning to adapt in dynamic, real-world environments through
  meta-reinforcement learning.
\newblock In \emph{ICLR}, 2018.

\bibitem[Tan et~al.(2018)Tan, Zhang, Coumans, Iscen, Bai, Hafner, Bohez, and
  Vanhoucke]{tan_2018}
J.~Tan, T.~Zhang, E.~Coumans, A.~Iscen, Y.~Bai, D.~Hafner, S.~Bohez, and
  V.~Vanhoucke.
\newblock Sim-to-real: Learning agile locomotion for quadruped robots.
\newblock In \emph{Robotics: Science and Systems}, 2018.
\newblock URL \url{https://arxiv.org/pdf/1804.10332.pdf}.

\bibitem[Mitchell et~al.(2020)Mitchell, Blumenkamp, and
  Prorok]{mitchell_2020_gp}
R.~Mitchell, J.~Blumenkamp, and A.~Prorok.
\newblock Gaussian process based message filtering for robust multi-agent
  cooperation in the presence of adversarial communication.
\newblock \emph{CoRR}, abs/2012.00508, 2020.
\newblock URL \url{https://arxiv.org/abs/2012.00508}.

\bibitem[Calvo-Fullana et~al.(2021)Calvo-Fullana, Mox, Pyattaev, Fink, Kumar,
  and Ribeiro]{calvo_2021_rosnetsim}
M.~Calvo-Fullana, D.~Mox, A.~Pyattaev, J.~Fink, V.~Kumar, and A.~Ribeiro.
\newblock Ros-netsim: A framework for the integration of robotic and network
  simulators.
\newblock \emph{IEEE Robotics and Automation Letters}, 6\penalty0 (2):\penalty0
  1120--1127, 2021.

\bibitem[Zhang et~al.(2019)Zhang, Tai, Yun, Xiong, Liu, Boedecker, and
  Burgard]{zhang_2019_vr_goggles}
J.~Zhang, L.~Tai, P.~Yun, Y.~Xiong, M.~Liu, J.~Boedecker, and W.~Burgard.
\newblock Vr-goggles for robots: Real-to-sim domain adaptation for visual
  control.
\newblock \emph{IEEE Robotics and Automation Letters}, 4\penalty0 (2):\penalty0
  1148--1155, 2019.

\bibitem[Mitchell et~al.(2020)Mitchell, Fletcher, Panerati, and
  Prorok]{mitchell_2020_mixed_reality}
R.~Mitchell, J.~Fletcher, J.~Panerati, and A.~Prorok.
\newblock Multi-vehicle mixed reality reinforcement learning for autonomous
  multi-lane driving.
\newblock In \emph{Proceedings of the 19th International Conference on
  Autonomous Agents and MultiAgent Systems}, AAMAS '20, page 1928–1930,
  Richland, SC, 2020. International Foundation for Autonomous Agents and
  Multiagent Systems.
\newblock ISBN 9781450375184.

\bibitem[McMahan et~al.(2017)McMahan, Moore, Ramage, Hampson, and
  y~Arcas]{mcmahan_2017_federated}
B.~McMahan, E.~Moore, D.~Ramage, S.~Hampson, and B.~A. y~Arcas.
\newblock Communication-efficient learning of deep networks from decentralized
  data.
\newblock In \emph{Artificial intelligence and statistics}, pages 1273--1282.
  PMLR, 2017.

\bibitem[Wang et~al.(2021)Wang, Xie, and Atanasov]{wang_2021_coding_federated}
B.~Wang, J.~Xie, and N.~Atanasov.
\newblock Coding for distributed multi-agent reinforcement learning.
\newblock \emph{arXiv preprint arXiv:2101.02308}, 2021.

\bibitem[Le et~al.(2017)Le, Yue, Carr, and Lucey]{le2017coordinated}
H.~M. Le, Y.~Yue, P.~Carr, and P.~Lucey.
\newblock Coordinated multi-agent imitation learning.
\newblock In \emph{International Conference on Machine Learning}, 2017.
\newblock URL \url{http://proceedings.mlr.press/v70/le17a/le17a.pdf}.

\bibitem[Lowe et~al.(2017)Lowe, Wu, Tamar, Harb, Abbeel, and
  Mordatch]{lowe_MultiAgent_2017a}
R.~Lowe, Y.~Wu, A.~Tamar, J.~Harb, P.~Abbeel, and I.~Mordatch.
\newblock Multi-{Agent} {Actor}-{Critic} for {Mixed}
  {Cooperative}-{Competitive} {Environments}.
\newblock In \emph{Proceedings of the 31st International Conference on Neural
  Information Processing Systems}, NIPS'17, page 6382–6393, Red Hook, NY,
  USA, 2017. Curran Associates Inc.
\newblock ISBN 9781510860964.

\bibitem[Damani et~al.(2021)Damani, Luo, Wenzel, and
  Sartoretti]{Damani2021_PRIMAL}
M.~Damani, Z.~Luo, E.~Wenzel, and G.~Sartoretti.
\newblock Primal$_2$: Pathfinding via reinforcement and imitation multi-agent
  learning - lifelong.
\newblock \emph{IEEE Robotics and Automation Letters}, 6\penalty0 (2):\penalty0
  2666--2673, 2021.

\end{thebibliography}

\end{document}